\documentclass[runningheads]{llncs}

\usepackage{graphicx}
\usepackage{float}
\usepackage{amsmath}
\usepackage{amssymb}
\usepackage{hyperref}
\usepackage{makecell}
\usepackage{subcaption}
\usepackage[export]{adjustbox}
\usepackage{array}
\usepackage{xcolor}

\definecolor{agentred}{RGB}{0, 0, 0}
\definecolor{agentpurple}{RGB}{0,0,0}
\definecolor{arrowgreen}{RGB}{0,0,0}

\begin{document}
\title{Theory of Mind Using Active Inference: \\ 
A Framework for Multi-Agent Cooperation}
\titlerunning{Theory of Mind Using Active Inference}


\author{Riddhi J. Pitliya\inst{1} \and 
Ozan Çatal\inst{1} \and 
Toon Van de Maele\inst{1} \and
Corrado Pezzato\inst{1} \and
Tim Verbelen\inst{1}
}

\authorrunning{Pitliya et al.}

\institute{VERSES, Los Angeles, California, CA 90067, USA \\
\email{riddhi.jain@verses.ai}}

\maketitle 

\begin{abstract}

Theory of Mind (ToM) -- the ability to understand that others can have differing knowledge and goals -- enables agents to reason about others' beliefs while planning their own actions. We present a novel approach to multi-agent cooperation by implementing ToM within active inference. Unlike previous active inference approaches to multi-agent cooperation, our method neither relies on task-specific shared generative models nor requires explicit communication. In our framework, ToM-equipped agents maintain distinct representations of their own and others' beliefs and goals. ToM agents then use an extended and adapted version of the sophisticated inference tree-based planning algorithm to systematically explore joint policy spaces through recursive reasoning. We evaluate our approach through collision avoidance and foraging simulations. Results suggest that ToM agents cooperate better compared to non-ToM counterparts by being able to avoid collisions and reduce redundant efforts. Crucially, ToM agents accomplish this by inferring others' beliefs solely from observable behaviour and considering them when planning their own actions. Our approach shows potential for generalisable and scalable multi-agent systems while providing computational insights into ToM mechanisms.


\end{abstract}

\section{Introduction}

Theory of Mind (ToM) represents one of the most remarkable achievements of human cognition -- the ability to understand that other agents possess minds with beliefs and goals that may differ from our own~\cite{frith2005theory}. This meta-cognitive skill enables us to recognise that others may hold false beliefs and maintain perspectives distinct from our own. For example, when we observe someone searching for an object that we know has been moved in their absence, we can anticipate their search behaviour based on where they believe the object to be, rather than its actual location~\cite{baron1985does,wimmer1983beliefs}. This fundamental distinction between reality and belief enables sophisticated forms of cooperation, competition, and communication. ToM emerges early in human development and underpins our ability to navigate complex multi-agent environments~\cite{wellman2001meta}. 

While ToM is fundamental to human social cognition, current approaches to multi-agent cooperation using active inference lack this crucial capability. Previous active inference models for multi-agent cooperation have predominantly relied upon assumptions of shared or identical generative models that limit their generalisability and practical application. We propose that a better solution to conduct and model multi-agent cooperation is by implementing ToM within the planning stage of active inference. This offers a more principled and generalisable solution for multi-agent artificial intelligent systems, and a computational model that could serve as a tool to deepen our understanding of how humans implement ToM. Before presenting our novel approach, we first elaborate on the limitations of existing approaches of conducting cooperation using active inference.

\subsection{Existing Active Inference Approaches to Multi-Agent Cooperation}

Maisto and colleagues~\cite{maisto2023interactive} introduced ``interactive inference'', wherein agents maintain probabilistic beliefs about shared goals (such as both agents pressing the same or different buttons) and update these beliefs through observations of others' locations and actions. Their agents selected epistemic policies designed to reduce uncertainty about the joint goals. However, this approach assumes agents share identical goals, which is not always the case in multi-agent cooperation tasks. Moreover, their model relies on a carefully tailored generative model for the focal agent (i.e., the agent conducting ToM) that incorporates the other agent's location as an observation which itself encodes information about the shared goal. These assumptions limit generalisability to scenarios where actions do not inherently signal goals or where agents have complementary rather than identical and shared objectives.

Matsumura and colleagues~\cite{matsumura2024active} addressed collision avoidance (agents passing by each other without colliding) using simulation theory, where agents use their own internal models to imagine others' situations. While this enables basic perspective-taking, their implementation is domain-specific to navigation tasks that use the social force model as it includes parameters for forward movement and mutual repulsion. The approach lacks the recursive reasoning capabilities that relies on maintenance of separate belief representations for different agents, which is necessary for more complex coordination scenarios.

Other researchers have proposed multi-agent cooperation through explicit information exchange mechanisms~\cite{catal2024belief,friston2024federated}. These approaches involve agents sharing likelihood messages -- information about the probability of observations given states -- rather than posterior beliefs directly. However, this requires the generative model structures (state factors and observation modalities) to be identical between agents. Moreover, while mathematically principled, this method sidesteps the fundamental challenge of inferring others' beliefs from observable behaviour alone, a capability that humans routinely demonstrate and leverage during multi-agent cooperation.

Overall, these approaches predominantly assume that all agents operate under the same generative model, with identical beliefs about transition dynamics, observation likelihoods and goal structures. Such aligned models fail to capture the reality of agents with different experiences, capabilities, and objectives. Furthermore, these approaches typically involve only single-level reasoning (\textit{``what will the other agent do?''}) rather than the recursive beliefs that characterise ToM (\textit{``what do I think the other agent thinks about the situation?''}). Many implementations are also tailored to specific tasks (e.g., navigation or mutual button-pressing) and do not provide general principles for multi-agent cooperation across different tasks.

To address these limitations, we present the first generalisable implementation for multi-agent cooperation by implementing ToM using sophisticated active inference~\cite{friston2021sophisticated}, with three key features:
\begin{enumerate}
    \item Our agents maintain distinct beliefs and generative models for themselves and others, allowing them to reason about different perspectives while avoiding the assumption of shared knowledge and knowledge structures.
    \item We propose a novel tree-based planning procedure that systematically explores joint policy spaces by interleaving policy and observation rollouts between agents.
    \item Our agents can reason about how others' actions affect the world states via message passing between their own beliefs and their beliefs of the other agent's beliefs, maintaining perspective separation while allowing for information integration.
\end{enumerate}
We empirically validate our approach by simulating two multi-agent scenarios: a collision avoidance task where agents must navigate past each other without occupying the same location, and an apple foraging task requiring efficient search and consumption of resources. These scenarios are implemented in a simple 3×3 grid environment to provide a clear and interpretable proof of concept, with future work aimed at extending the approach to larger and more complex settings. Our results demonstrate that ToM-equipped agents conduct multi-agent cooperation more effectively than non-ToM agents. The ToM agents successfully avoid collisions and reduce redundant efforts as they are able to interact \textit{proactively} rather than reactively with other agents. 

\section{Our Approach: Theory of Mind in Active Inference}

\subsection{Sophisticated Inference}

Our approach builds upon sophisticated inference~\cite{friston2021sophisticated}, which extends standard active inference to consider recursive forms of expected free energy (EFE). In standard active inference, agents evaluate policies by considering \textit{``what would happen if I did that?''}. Sophisticated inference instead deepens this to \textit{``what would I believe about what would happen if I did that?''}. This distinction is crucial for ToM. When reasoning about other agents, we need to consider not just what they will do, but what they believe about the consequences of their actions. This recursive reasoning about the other agent requires maintaining a separate model of the other agent. 

\subsection{ToM Agent's Belief Structure of Multiple Agents}
\label{sec:multibeliefs}

In our ToM framework, the focal agent maintains separate state beliefs ($s$) for itself and each other agent in the environment. In the case of a two-agent scenario, this yields $s = \{\underbrace{s^{f,\text{self}}, s^{f,\text{world}}}_{\text{focal ($s^{f}$)}}, \underbrace{s^{o,\text{self}}, s^{o,\text{world}}}_{\text{other ($s^{o}$)}}\}$, where:

\begin{itemize}
    \item $s^{f,\text{self}}$ is the focal agent's beliefs about its own states (e.g., focal agent's own location)
    \item $s^{f,\text{world}}$ is the focal agent's beliefs about the world states (e.g., other agent's location or item at current location)
    \item $s^{o,\text{self}}$ is the focal agent's beliefs about the other agent's self states (e.g., other agent's location)
    \item $s^{o,\text{world}}$ is the focal agent's beliefs about what the other agent believes about the world states (e.g., focal agent's location or item at current location)
\end{itemize}

This structure lets the focal agent keep its own perspective separate while modelling how others might see things differently. To elaborate, the belief components can be flexibly combined to capture different reasoning cases. For instance, the focal agent can pair $s^{o,\text{self}}$ with $s^{f,\text{world}}$ to predict what the other agent would observe given the focal agent's own beliefs about the environment. The focal agent can also pair $s^{o,\text{self}}$ with $s^{o,\text{world}}$ to predict what the other agent thinks it will observe, based on the other agent’s own (possibly mistaken) view of the world. This kind of cross-perspective reasoning allows the focal agent to distinguish between (a) what it thinks the other agent will perceive and (b) what it thinks the other agent believes it may perceive. Because the focal agent's own world beliefs ($s^{f,\text{world}}$) may differ from its beliefs about the other agent’s world beliefs ($s^{o,\text{world}}$), it can represent cases where knowledge is asymmetric -- for example, when the focal agent knows something the other does not.

By maintaining these distinct representations, our framework does not assume shared knowledge structures between agents. The focal agent can construct and continuously update its model of the other agent based on observable behaviour, while the actual other agent may operate with an entirely different generative model. This capability enables our agents to collaborate effectively even when they possess different prior knowledge, capabilities, or goals -- a fundamental requirement for realistic and practical multi-agent systems.

\begin{figure}[t]
    \centering
    \includegraphics[width=\textwidth]{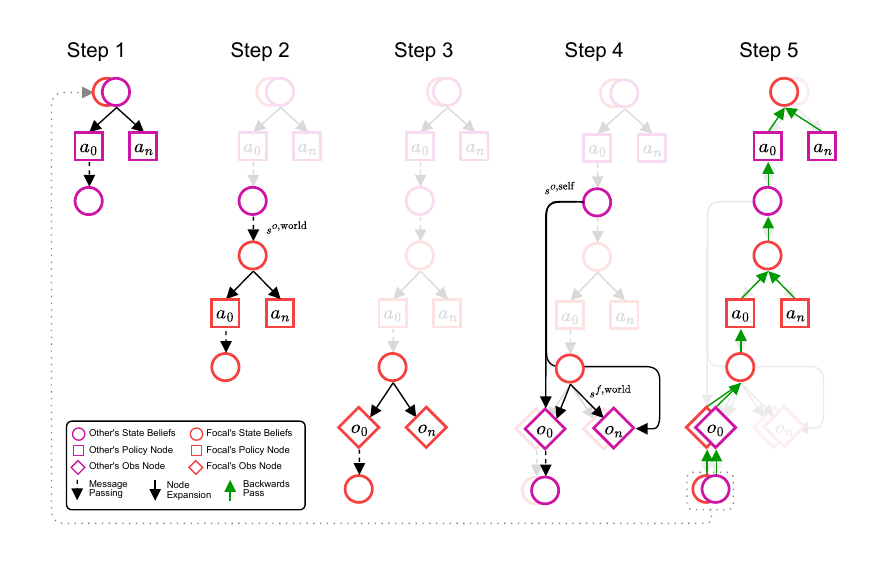}
    \caption{\textbf{Recursive Planning Tree for Theory of Mind.} \textcolor{agentred}{Red} and \textcolor{agentpurple}{purple} represent the focal and other agent's nodes respectively. Circles indicate the agent's beliefs, squares indicate evaluated actions and diamonds indicated expected observations. For a detailed description of each step, see Section~\ref{sec:planningtree}.}
    \label{fig:planningtree}
\end{figure}

\subsection{Recursive Planning with Theory of Mind}
\label{sec:planningtree}

The core innovation in our approach lies in the planning algorithm that enables agents to reason recursively about joint policy spaces. It systematically explores how the focal agent's beliefs about another agent's beliefs influence its planning decisions. The recursive form of the EFE is extended to the ToM setting (see Appendix~\ref{app:sophisticated}), which results in a deep tree search algorithm that alternates between the focal and other agent's policies and observations. At each planning horizon, the tree search unfolds through five main stages, as detailed below and illustrated in Figure~\ref{fig:planningtree}.

\textbf{Step 1: Other Agent Policy Expansion.}
As aforementioned in Section~\ref{sec:multibeliefs}, we begin with the focal agent's beliefs, which comprises separate beliefs for itself and the other agent in the environment ($s=\{s^{f}, s^{o}\}$). The focal agent first considers which policies the other agent is likely to select. This is visualised in Step 1 of Figure~\ref{fig:planningtree}, where each policy node represents a specific action that the other agent might execute ($a_0$; \textcolor{agentpurple}{purple square}). The potential actions are evaluated based on the focal agent's beliefs about the other agent's beliefs ($s^{o}$; \textcolor{agentpurple}{purple circle} above \textcolor{agentpurple}{purple squares}). Essentially, the focal agent asks \textit{``Given what I believe about the other agent's beliefs and goals, what would the other agent choose to do?''}. The focal agent then computes how the other agent would update its beliefs if it were to execute that action ($s^{o}$; the \textcolor{agentpurple}{purple circle} below the \textcolor{agentpurple}{purple square}).

\textbf{Step 2: Focal Agent Policy Expansion.}
For each considered action of the other agent, the focal agent evaluates its own policy options. Crucially, before doing so, the focal agent updates its world beliefs, based on the anticipated consequences of the other agent's actions, using likelihood message passing. Here, the focal agent uses its computation of how the other agent's beliefs about the world ($s^{o,\text{world}}$; \textcolor{agentpurple}{purple circle} above the \textcolor{agentred}{red circle}) would change if the other agent were to execute an action. A likelihood message is then created, capturing the information gained from the other agent's anticipated action in the form of the difference between the other agent's updated beliefs and its prior beliefs. This mechanism allows the focal agent to incorporate information into its own beliefs about how the world states ($s^{f,\text{world}}$; \textcolor{agentred}{red circle} above the \textcolor{agentred}{red squares}) would change due to the other agent's actions. Using these updated beliefs, the focal agent then evaluates its own policy options ($a_0$; \textcolor{agentred}{red square}) through standard EFE calculations. This creates branches in the tree structure for each possible joint policy combination between the focal and other agent. The focal agent then computes how its beliefs would be updated if it were to execute an action given the other agent's action ($s^{f}$; \textcolor{agentred}{red circle} below the \textcolor{agentred}{red square}).

\textbf{Step 3: Focal Agent Observation Expansion.}
Then, given the joint policies, the focal agent considers what observations it is likely to receive and its resulting posterior beliefs.

This process is illustrated in Step 3 of Figure~\ref{fig:planningtree}, where the focal agent's observation nodes ($o_0$; \textcolor{agentred}{red diamonds}) represent the various observations the focal agent expects to encounter given the focal agent's beliefs considering the execution of both agents' actions ($s^{f}$; \textcolor{agentred}{red circle} before \textcolor{agentred}{red diamonds}). This results in the computation of the focal agent's posterior beliefs ($s^{f}$; \textcolor{agentred}{red circle} after \textcolor{agentred}{red diamonds}).

\textbf{Step 4: Other Agent Observation Expansion.}
Here, the focal agent considers what observations the other agent is likely to receive ($o_0$; \textcolor{agentpurple}{purple diamonds}) given the joint policy and anticipated world state changes given its action. The observation probabilities are computed using the focal agent's beliefs about the other agent's self states ($s^{o,\text{self}}$; \textcolor{agentpurple}{purple circle} from an earlier expansion) and the focal agent's own updated beliefs about the world state ($s^{f,\text{world}}$; \textcolor{agentred}{red circle} before the \textcolor{agentpurple}{purple diamonds}). The focal agent then updates its representation of the other agent's posterior beliefs ($s^{o}$; \textcolor{agentpurple}{purple circle} after the \textcolor{agentpurple}{purple diamond}).

\textbf{Step 5: Tree Backwards Pass and Policy Selection.} Finally, after expanding the tree for the current horizon, a backwards pass computes policy selection probabilities for the focal agent. The backwards pass is visualised in Step 5 of Figure~\ref{fig:planningtree}, with the \textcolor{arrowgreen}{green upward arrows} showing that EFE values propagate from the leaf observation nodes back through each policy branch to inform the final policy selection at the root. To plan for another time step, the observation nodes' leaves from Step 5 serve as the root node for Step 1 (\textcolor{black}{grey dotted arrow}).

Recursive EFE values are computed for each joint policy combination and weighted by the observation probabilities. The other agent's policy probabilities are marginalised for policy selection. The resulting probability distribution balances goal-directed with information-seeking behaviour while taking into account the uncertainty over the other agent's actions.

Our implementation achieves computational efficiency through two mechanisms as practised in sophisticated active inference~\cite{friston2021sophisticated}. Policy pruning reduces tree expansion by eliminating unlikely policy nodes and those nodes do not branch out. Observation pruning similarly focuses on probable outcomes, reducing the combinatorial explosion.

\section{Experimental Validation}

We empirically validated our ToM framework across two multi-agent scenarios that required different forms of cooperation. All simulations occurred on a 3×3 grid environment (see Appendix~\ref{app:treevisualisations_collisionavoidance} for the reference grid layout) with deterministic dynamics and perfect observability of agent locations. The experimental design comprised two conditions for each task: a baseline condition where both agents used sophisticated active inference without ToM capabilities, and a ToM condition where one agent (red) was equipped with our ToM framework while the other (purple) remained non-ToM. All simulations were conducted using the JAX-based Python package, \texttt{pymdp}, which offers efficient and flexible tools for constructing such models \cite{heins2022pymdp}.

\subsection{Collision Avoidance Task}

\subsubsection{Task Description.}

The collision avoidance task presented a basic cooperation challenge: two agents were initiated at opposite corners of the grid with objectives to swap positions while avoiding collision. The shortest path for both agents involved traversing the central cell that would result in collision. We evaluated performance using three primary metrics: task completion success (whether agents reached their respective goals), collision occurrence, and path efficiency (total time steps to completion). The task demanded proactive cooperation, as reactive strategies would result in deadlock.

\subsubsection{Generative Model.}

Each agent's generative model included two state factors: own location (9 discrete location states plus a null state for boundary violations as in~\cite{van2023integrating}) and other agent's location (also 10 states like the focal agent's location states). The observation model provided perfect perceptual access to both locations through identity likelihood mappings, eliminating sensory uncertainty.

The action space comprised nine options: directional movements (up, down, left, right, four diagonals) plus no-operation. Transition dynamics regarding the agent's own location reflected full controllability abiding by standard grid-world physics with collision constraints where agents attempting to occupy the same cell became permanently stuck. Invalid movements were specified via the null state (which had a severe negative utility). For example, if the agent were to move up from location 1 (top-left corner), it would enter the severely disfavoured null state, driving it away from invalid movements. Since the other agent's location was not controllable, the transition dynamics for the other agent's location reflected uniform probability distribution between the valid actions the other agent can take. For example, the probability of moving from location 1 (top-left corner) to the centre, go down, go right, or have no-operation is 1/4.

Preferences encoded goal-seeking behaviour: high positive utility for reaching the target location and severe penalty for the null observation. Critically, no explicit collision avoidance preferences were included -- coordination had to emerge through ToM reasoning rather than hard-coded behaviours. Planning horizons were set to 3 time steps for both ToM and non-ToM agents, sufficient to reach goals via alternative paths while requiring that much forward planning to identify coordination opportunities.

\begin{figure}[t!]
    \centering
    \subcaptionbox[]{\label{fig:collisionavoidance}}{
        \includegraphics[width=0.525\textwidth]{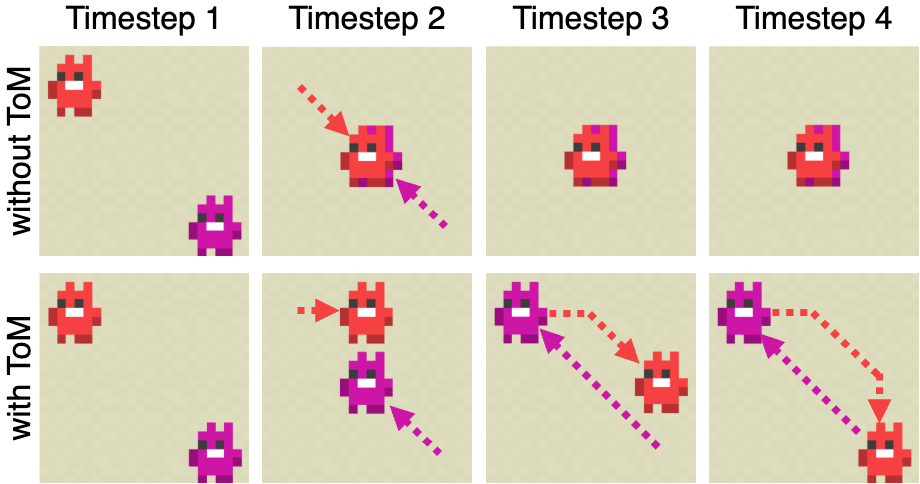}
    }
    \hfill
    \subcaptionbox[]{\label{fig:appleforaging}}{
        \includegraphics[width=0.4\textwidth]{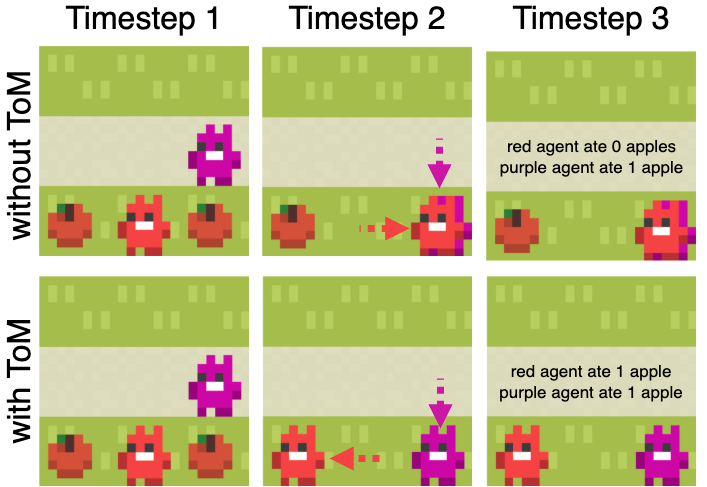}
    }
    \caption{\textbf{Experimental Validation of Theory of Mind in Active Inference Using Multi-Agent Cooperation Tasks.} 
    (a) Collision avoidance task: agents must swap locations while avoiding collision at the central location. 
    \textit{Top row}: Non-ToM condition where both agents select the shortest path, resulting in collision and deadlock. 
    \textit{Bottom row}: ToM condition where the red agent (equipped with ToM) anticipates the purple agent's path and selects an alternative route, enabling successful cooperation. 
    (b) Apple foraging task: agents search for and consume apples in orchard locations, with both agents initially knowing an apple exists at the bottom-right corner. 
    \textit{Top row}: Non-ToM condition showing resource competition where both agents converge on the known apple location, resulting in only one agent (purple) successfully consuming the apple. 
    \textit{Bottom row}: ToM condition where the red agent (equipped with ToM) explores an alternative location to avoid redundant competition, successfully finding and consuming an apple while the purple agent consumes the known apple, achieving optimal resource allocation for both agents.}
    \label{fig:tom_experiments}
\end{figure}

\subsubsection{Results.}

For an illustration of the results from this task, see Figure \ref{fig:collisionavoidance}. In the non-ToM condition, both agents predictably selected individually optimal policies, moving directly toward goals via the centre. This resulted in collision and permanent deadlock, with neither agent achieving its objective. This is a clear cooperation failure despite sophisticated individual planning while observing the other agent's location but not being equipped to incorporate it into the planning process.

In the ToM condition, the red (ToM) agent reasoned that the other agent will most likely move to the central location to take the shortest path to its goal, so it does not select moving to that location to avoid collision even though it was the most optimal policy for itself. Instead, the agent went around the centre, which was the next best alternative route towards its goal. The ToM agent selected a longer but collision-free route. For the detailed planning tree of the ToM vs non-ToM agents at time step 0, see Appendix~\ref{app:treevisualisations_collisionavoidance}.

\subsection{Apple Foraging Task}

\subsubsection{Task Description.}

The apple foraging task examined cooperation in resource acquisition scenarios with partial observability. The 3×3 grid featured orchard locations (top and bottom rows) where apples could spawn, and wasteland (middle row) containing no resources. Both agents began with identical prior knowledge: certainty of an apple at the bottom-right corner, with complete uncertainty about the presence of an apple at other orchard locations. The agents' initial positions were equidistant from the known apple (see Figure~\ref{fig:appleforaging}). Apple consumption was exclusive -- only one agent could consume each apple, with random selection if both agents reached the same apple simultaneously. The cooperation challenge involved balancing exploitation of known resources against exploration of uncertain locations, while avoiding redundant competition for the same resources.

\subsubsection{Generative Model.}

The generative model incorporated three types of state factors: agent locations, reward feedback (binary: received/not received, which was conditioned upon eating an apple), and environmental items (wasteland, apple, or empty orchard). Agents observed their own location, the other agent's location, items at their current location, and their own reward feedback.

The environment was partially observable such that agents could only assess apple availability at their current location, creating uncertainty about the broader resource distribution. Apples spawned probabilistically in orchard locations (25\% per time step) and remained in that location until consumption. Action repertoires included movements (up, down, left, right actions), eating, and no operation. Preferences simply favoured reward acquisition without explicit cooperation incentives. Planning horizons were set to 3 time steps, sufficient for reaching the opposing end of the grid environment and consuming apples.

\subsubsection{Results.}

For an illustration of the results from this task, see Figure~\ref{fig:appleforaging}. For the detailed planning tree of the ToM vs non-ToM agents at time step 0, see Appendix~\ref{app:treevisualisations_foraging}.

In the non-ToM condition, both agents converged on the known apple location (bottom-right corner), resulting in resource competition. Only one agent succeeded in consuming the apple (determined randomly), while the other wasted effort, demonstrating inefficient cooperation.

In the ToM condition, the red (ToM) agent reasoned that the other agent will most likely go to the known apple location and therefore, it selected to explore another location where it was uncertain whether there is an apple. This resulted in avoidance of redundant efforts and more effective cooperation, avoiding resource competition. This strategy proved successful as both agents discovered and consumed apples.

\section{Discussion}
Our experimental results demonstrate that equipping active inference agents with ToM capabilities fundamentally transforms their approach to completing a task, conducting multi-agent cooperation. The ToM-equipped agents successfully navigated both collision avoidance and resource competition scenarios, achieving better cooperation compared to their non-ToM counterparts. Importantly, this enhanced performance emerged without requiring explicit communication protocols, shared generative models between agents, or pre-set strategies.

The success of our ToM agents stems from their ability to reason about others' beliefs and anticipate their behaviours. In the collision avoidance task, the ToM agent recognised that both agents following optimal individual paths would result in collision. By reasoning about the other agent's likely trajectory, the ToM agent proactively selected the next best alternative route, demonstrating cooperative behaviour compared to a merely reactive collision response. Similarly, in the apple foraging task, the ToM agent anticipated resource competition and strategically explored uncertain locations, leading to more efficient resource allocation across both agents.

Our framework addresses fundamental limitations from previous multi-agent active inference implementations. Most importantly, we eliminate the restrictive assumption of shared or identical generative models that dominates prior work \cite{maisto2023interactive,matsumura2024active,catal2024belief,friston2024federated}. This limited their generalisability and applicability to more complex or real-world scenarios where agents possess different experiences, capabilities, and objectives. In contrast, our ToM framework allows for heterogeneous multi-agent generative models. The ToM agent maintains distinct belief representations for each agent in its environment, allowing it to reason about others without assuming they share its own knowledge, goals, or even generative model structure. 

\subsection{Future Directions}

While this paper provides valuable insights into computationally conducting multi-agent cooperation by implementing ToM in active inference, there are several avenues for future research to build upon our findings.

Our current implementation assumes observational access to the other agents' locations and is situated in a simple 3x3 grid environment. While these simplifications enables clear demonstration of the core ToM principles, we naturally need to examine it under more complex and real-world scenarios as such scenarios involving more noisy sensory information and complex task requirements and dynamics. Future work should also include systematic quantitative evaluation using aggregated performance metrics across random seeds and statistical comparisons against non-ToM baselines to more rigorously assess the robustness of our approach.

Another simplification in the reported simulations was that our ToM agents assumed knowledge of others' goals and operated with fixed generative models of other agents. Future implementations should incorporate online learning mechanisms, potentially using Dirichlet counts \cite{friston2016active}, to continuously learn and update beliefs about the other agents' model, preferences, and capabilities. Such adaptive learning would significantly enhance the framework's generalisability, enabling effective cooperation with agents whose characteristics are initially unknown or evolving.

Moreover, while our current implementation focuses on dyadic interactions, the underlying principles naturally extend to larger multi-agent scenarios. Each agent would maintain separate belief representations for all other agents, and the planning algorithm would expand over joint policy spaces of arbitrary size. However, computational complexity grows exponentially with the number of agents, presenting scalability challenges that require careful considerations and is an avenue of further research. 

Furthermore, our implementation focuses on first-order ToM reasoning (\textit{``what does the other agent believe?''}) rather than higher-order recursive reasoning (\textit{''what do I think the other agent thinks I believe?''}). While first-order ToM proves to be sufficient for our tested cooperation scenarios, more complex social situations may require deeper levels of recursive reasoning. This could be examined in scenarios where there are multiple ToM agents, investigating how recursive reasoning between ToM-capable agents affects cooperation dynamics and computational requirements.

Additionally, our framework has been validated only in cooperative scenarios where agents' goals are somewhat complementary. Future research should investigate competitive scenarios, where agents' objectives directly conflict, to assess whether the planning algorithm remains effective and how generative models should be structured to handle adversarial interactions.

\section{Conclusion}

We have presented the first generalisable implementation of ToM within the active inference framework for multi-agent cooperation. Our approach represents a significant advancement over existing methods by eliminating the restrictive assumption that all agents must operate under shared or identical generative models and goal structures.

The core innovation of our framework lies in enabling agents to maintain distinct beliefs about themselves and others while reasoning recursively about how others' beliefs influence their behaviour. Through our novel tree-based planning algorithm, ToM-equipped agents systematically explore joint policy spaces by considering what others believe and how those beliefs influence their consideration and decisions of actions. This recursive reasoning capability allows for sophisticated cooperation online without requiring explicit communication or pre-arranged cooperation protocols.

We validated our framework with two tasks: collision avoidance and resource foraging tasks. ToM agents successfully cooperated in both scenarios, avoiding conflicts and achieving more efficient outcomes compared to non-ToM agents. Importantly, these cooperative capabilities emerged immediately upon encountering the task challenges, without requiring lengthy training periods or domain-specific learning.

The framework we have developed bridges computational and cognitive science, providing both a practical tool for enhancing artificial intelligence systems and a computational foundation for understanding how sophisticated social reasoning might emerge from principled probabilistic inference about others' minds.

 
\bibliography{refs}
\bibliographystyle{splncs04}

\clearpage

\appendix

\section{Expected Free Energy under Sophisticated Inference with Theory of Mind}
\label{app:sophisticated}

In sophisticated inference, the expected free energy is calculated in a recursive way~\cite{friston2021sophisticated}. This can be construed as a deep tree search, where the tree branches over allowable actions at each point in time, and the likely observations consequent upon each action. Equation~\ref{eq:si_efe} expresses the expected free energy of each potential next action ($a_\tau$) and observation ($o_\tau$) as the
utility and information gain of that action plus the average expected free energy of future beliefs, under counterfactual observations and actions:

\begin{align}
    G(o_{\tau}, a_{\tau}) &= \mathbb{E}_{Q(o_{\tau+1} | a_{\le \tau})}\bigl[\underbrace{-\ln P(o_{\tau+1} | C )}_{\text{utility}} - \underbrace{D_{KL}[Q(s_{\tau+1} | o_{\tau+1} 
    ) || Q(s_{\tau+1} 
    )
    ]}_{\text{information gain}} \bigl]  \notag \\
    &  + \underbrace{\mathbb{E}_{Q(a_{\tau+1} | o_{\tau + 1})Q(o_{\tau+1} | a_{\le \tau})}\bigl[ 
    G(o_{\tau+1}, a_{\tau+1})}_{\text{expected free energy of subsequent actions}}
    \bigl]
\label{eq:si_efe}
\end{align}

with 

\begin{align}
Q(o_{\tau+1} | a_{\le \tau}) &= P(o_{\tau+1} | s_{\tau+1}) Q(s_{\tau+1} | a_{\le \tau}) \notag \\
Q(a_{\tau} | o_{\tau}) &= \sigma \bigl( -G(o_\tau, a_\tau) \bigl)  \notag
\end{align}

When planning with Theory of Mind, this deep tree search becomes an alternation between actions of the other agent and actions of the focal agent, as well as counterfactual observations for both. The resulting expected free energy in Equation~\ref{eq:tom_efe} is now expressed as a function of an action and observation for the focal agent ($a_\tau^f, o_\tau^f$) as well as the other agent ($a_\tau^o, o_\tau^o$):

\begin{align}
    G(o^{f}_{\tau},o^{o}_{\tau}, a^{f}_{\tau},  a^{o}_{\tau}) &= \mathbb{E}_{Q(o^{f}_{\tau+1}, o^{o}_{\tau+1} | a^{f}_{\le \tau}, a^{o}_{\le \tau})} \notag \\
    &\quad\quad\quad\bigl[-\ln P(o^{f}_{\tau+1} | C^{f})
    - \mathbb{D}_{\text{KL}}[Q(s^{f}_{\tau+1} | o^{f}_{\tau+1} 
    ) || Q(s^{f}_{\tau+1} 
    )
    ] \bigl]  \notag \\
    &\quad+ \mathbb{E}_{Q(a^{f}_{\tau+1} | o^{f}_{\tau + 1})Q(a^{o}_{\tau+1} | o^{o}_{\tau + 1})Q(o^{f}_{\tau+1}, o^{o}_{\tau+1} | a^{f}_{\le \tau}, a^{o}_{\le \tau})} \notag\\
    & \quad\quad\quad \bigl[ 
    G(o^{f}_{\tau+1}, o^{o}_{\tau+1}, a^{f}_{\tau+1}, a^{o}_{\tau+1})
    \bigl]
\label{eq:tom_efe}
\end{align}

with

\begin{align}
Q(o^{f}_{\tau+1}, o^{o}_{\tau+1} | a^{f}_{\le \tau}, a^{o}_{\le \tau}) &= P(o^{f}_{\tau + 1} | s^{f}_{\tau+1}) Q(s^{f}_{\tau + 1}) P(o^{o}_{\tau + 1} | s^{o}_{\tau+1}, s^{f}_{\tau+1}) Q(s^{o}_{\tau + 1}) \notag \\
Q(s^{f}_{\tau + 1}) &= Q(s^{f}_{\tau + 1} | s^{f}_{\tau}, a^{f}_{\tau}, s^{o}_{\tau+1})
\mathbb{E}_{Q(o^{o}_{\tau + 1} | a^{o}_{\le \tau})}[Q(s^{o}_{\tau + 1} | o^{o}_{\tau+1})] \notag \\
Q(a^o_{\tau} | o^o_{\tau}) &= \sigma \bigl( -G(o^o_\tau, a^o_{\tau}| C^o) \bigl)  \notag \\
Q(a^f_{\tau} | o^f_{\tau}) &= \sigma \bigl( -G(o^{f}_{\tau},o^{o}_{\tau}, a^{f}_{\tau},  a^{o}_{\tau}) \bigl)  \notag \\
\label{eq:tom_efe_with}
\end{align}

Note that the utility term for the focal agent uses the focal agent's preferences $C^f$, which might be distinct from the other agent's preferences $C^o$ which are used to calculate the posterior over the other's actions. In addition, our posterior over expected observations for the other agent is conditioned on both $s^{o}_{\tau+1}$ and $s^{f}_{\tau+1}$. Here, we combine the $s^{o,\text{self}}$ with $s^{f, \text{world}}$ to generate expected observations from the perspective of the other, using the world belief of the focal agent. Another crucial point in Equation~\ref{eq:tom_efe_with} is that to calculate the predictive posterior $Q(s^{f}_{\tau + 1})$ of the focal agent, we also incorporate the effect the other's action might already have had on the focal agent's belief state. This is implemented via belief sharing via likelihood message passing, similar to~\cite{catal2024belief}.


\section{Planning Trees for Collision Avoidance Task}
\label{app:treevisualisations_collisionavoidance}

\begin{figure}
    \centering
    \subcaptionbox[]{\label{fig:collision_avoided}}{
        \includegraphics[width=0.35\textwidth]{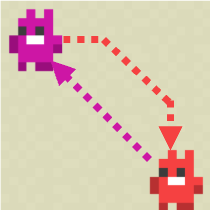}
    }
    \hfill
    \subcaptionbox[]{\label{fig:3by3}}{
        \includegraphics[width=0.35\textwidth]{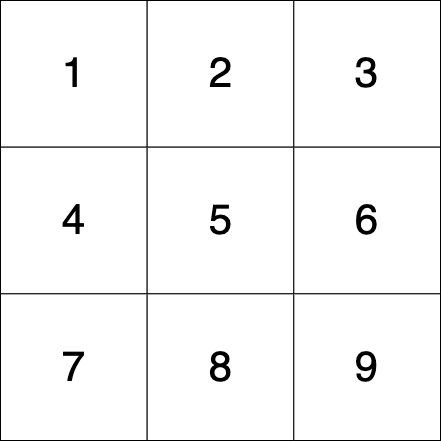}
    }
    
    \subcaptionbox[]{\label{fig:coll_treevis_nontom}}{
        \includegraphics[width=0.4\textwidth]{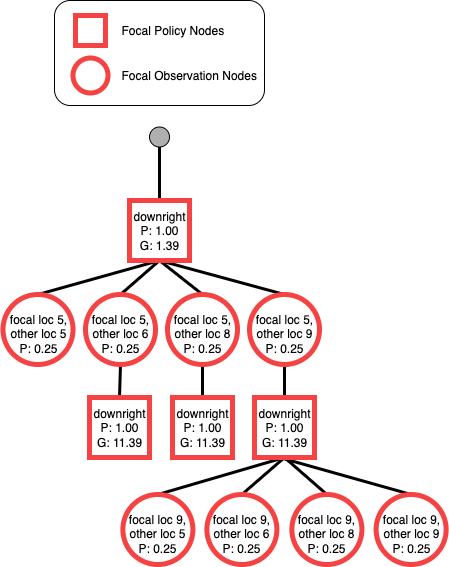}
    }
    \hfill
    \subcaptionbox[]{\label{fig:coll_treevis_tom}}{
        \includegraphics[width=0.3\textwidth]{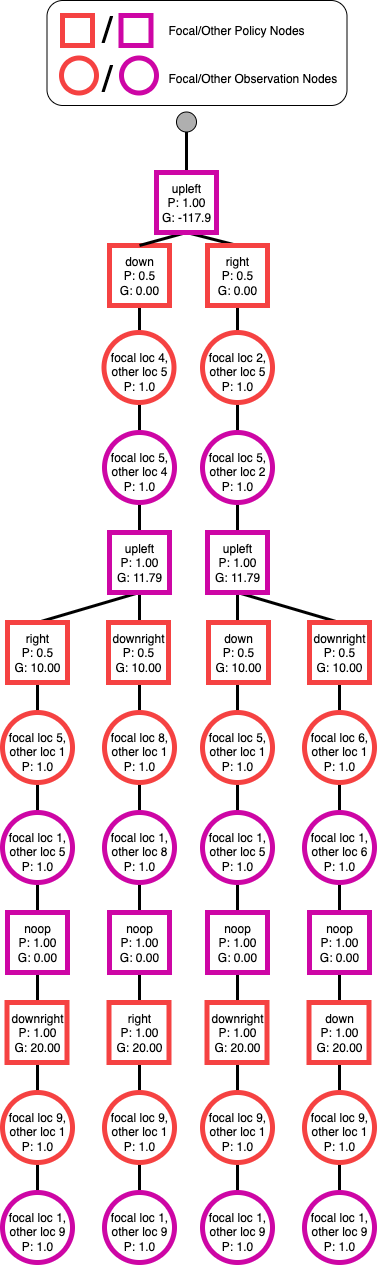}
    }
    
    \caption{(a) Collision avoided by the red agent as the focal agent with ToM. (b) Reference grid layout for 3×3 simulation environment. Numbers 1-9 indicate cell indices used throughout the experiments to specify agent locations and movements. 
    (c) Non-ToM planning tree: The red agent evaluates only its own policies over a 2-step horizon, selecting to go to location 5 (P$=$1.0), which eventually results in colliding with the other agent. (d) ToM planning tree: The red (ToM) agent recursively models the purple (non-ToM) agent's policy space and beliefs, resulting four possible routes the red agent can take to avoid collision with the other agent which is predicted to go to its goal location via location 5.}
    \label{fig:planning_trees_collisionavoidance}
\end{figure}

Figure~\ref{fig:planning_trees_collisionavoidance} shows the difference in planning trees for both a non-ToM (c) and ToM agent (d) completing the collision avoidance task with another agent. It is clear that the ToM agent's planning tree (d) entertains four routes the agent can take to avoid colliding with the other (non-ToM) agent and that the route illustrated in the right-most branch of the planning tree is selected to complete the task (a).  

\section{Planning Trees for Apple Foraging Task}
\label{app:treevisualisations_foraging}

\begin{figure}
    \centering
    \subcaptionbox[]{\label{fig:apple_foraging}}{
        \includegraphics[width=0.35\textwidth]{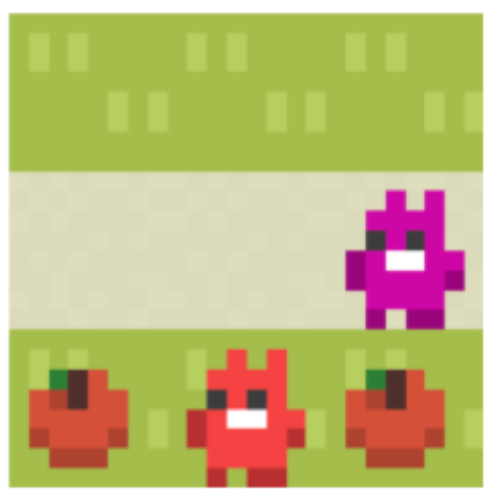}
    }
    \hfill
    \subcaptionbox[]{\label{fig:3by3b}}{
        \includegraphics[width=0.35\textwidth]{figures/3by3.png}
    }
    
    \subcaptionbox[]{\label{fig:foraging_treevis_nontom}}{
        \includegraphics[width=0.25\textwidth]{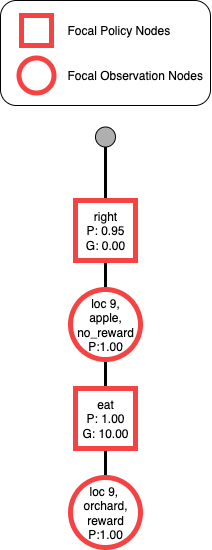}
    }
    \hfill
    \subcaptionbox[]{\label{fig:foraging_treevis_tom}}{
        \includegraphics[width=0.4\textwidth]{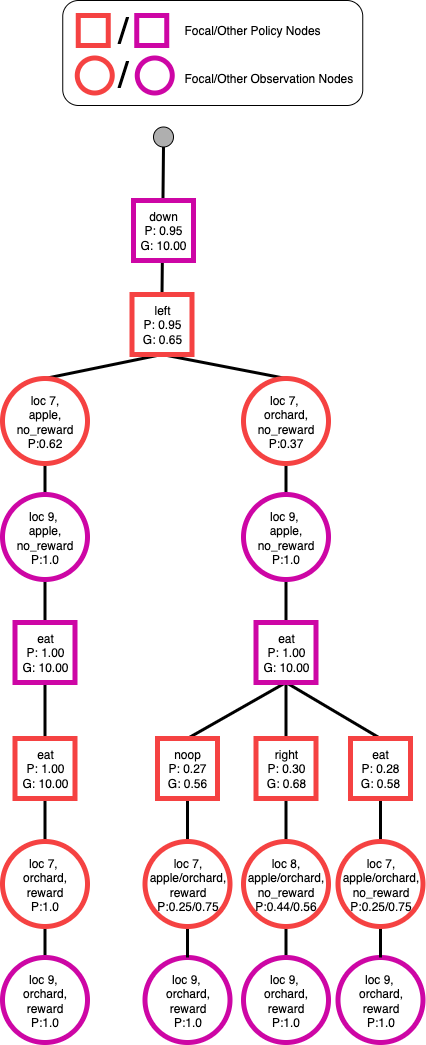}
    }
    
    \caption{(a) Start situation with the red agent as the focal agent. (b) Reference grid layout for 3×3 simulation environment. Numbers 1-9 indicate cell indices used throughout the experiments to specify agent locations and movements. (c) Non-ToM planning tree: The red agent evaluates only its own policies over a 2-step horizon, selecting to go to location 9 (P$=$1.0) based on expected utility (G$=$10.00). (d) ToM planning tree: The red (ToM) agent recursively models the purple (non-ToM) agent's policy space and beliefs, resulting in selection of going left with near full certainty (q\_pi$=$0.95). This recursive reasoning leads to exploration of location 7, as the agent predicts the purple agent will pursue location 9, enabling coordinated foraging.}
    \label{fig:planning_trees_foraging}
\end{figure}

Figure~\ref{fig:planning_trees_foraging} shows the difference in planning trees for both a non-ToM (c) and ToM agent (d) completing a apple foraging task with another agent. We start from the orchard environment state in (a), where both agents navigate to locate and consume apples, with initial shared knowledge of an apple at location 9. We visualise the planning tree for the red agent with (d) and without ToM (c).

In the non-ToM case, the focal agent (red) evaluates only its own policies over a 2-step horizon, selecting to go to location 9 (P$=$1.0) based on expected utility (G$=$10.00). However, when pursuing this policy it will end up with no apple, as the purple agent will get there first.

In contrast, in the ToM case, the focal agent (red) indeed reasons that the other agent (purple) will get to the apple first, and therefore chooses to explore the tile on the right, which might have an apple as well.

\end{document}